\pdfoutput=1
\documentclass{article}

    \usepackage[final,nonatbib]{neurips_2021}

\usepackage[utf8]{inputenc} 
\usepackage[T1]{fontenc}    
\usepackage{hyperref}       
\usepackage{url}            
\usepackage{booktabs}       
\usepackage{amsfonts}       
\usepackage{nicefrac}       
\usepackage{microtype}      
\usepackage{xcolor}         

\usepackage{multirow}
\usepackage{booktabs}
\usepackage{times}
\usepackage{epsfig}
\usepackage{graphicx}
\usepackage{amsmath}
\usepackage{amssymb}
\usepackage{algorithm}
\usepackage{algpseudocode}
\usepackage{xcolor}
\usepackage{float}

\title{How Far Can I Go ? : A Self-Supervised Approach for Deterministic Video Depth Forecasting}

\author{%
  Sauradip Nag\thanks{Equal Contribution} \thanks{Work done prior joining University of Surrey} \\
  CVSSP, University of Surrey, UK\\
  \texttt{s.nag@surrey.ac.uk} \\
   \And
  Nisarg A. Shah$^{*}$  \\
  IIT Jodhpur, India \\
  \texttt{shah.2@iitj.ac.in} 
   \AND
  Anran Qi$^{*}$ \\
  CVSSP, University of Surrey, UK \\
  \texttt{a.qi@surrey.ac.uk}
   \And
  Raghavendra Ramachandra \\
  NTNU, Norway \\
  \texttt{raghavendra.ramachandra@ntnu.no} 
}

\begin{document}

\maketitle

\begin{abstract}
  In this paper we present a novel self-supervised method to anticipate the depth estimate for a future, unobserved real-world urban scene. This work is the first to explore self-supervised learning for estimation of monocular depth of future unobserved frames of a video. 
    Existing works rely on a large number of annotated samples to generate the probabilistic prediction of depth for unseen frames.  However, this makes it unrealistic due to its requirement for large amount of annotated depth samples of video. In addition, the probabilistic nature of the case, where one past can have multiple future outcomes often leads to incorrect depth estimates. Unlike previous methods, we model the depth estimation of the unobserved frame as a view-synthesis problem, which treats the depth estimate of the unseen video frame as an auxiliary task while synthesizing back the views using learned pose. This approach is not only cost effective - we do not use any ground truth depth for training (hence practical) but also deterministic (a sequence of past frames map to an immediate future). To address this task we first develop a novel depth forecasting network DeFNet which estimates depth of unobserved future by forecasting latent features. Second, we develop a channel-attention based pose estimation network that estimates the pose of the unobserved frame. Using this learned pose, estimated depth map is reconstructed back into the image domain, thus forming a self-supervised solution. Our proposed approach shows significant improvements in Abs Rel metric compared to state-of-the-art alternatives on both short and mid-term forecasting setting, benchmarked on KITTI and Cityscapes. Code is available at \href{https://github.com/sauradip/depthForecasting}{https://github.com/sauradip/depthForecasting}
\end{abstract}

\section{Introduction}
\label{sec:intro}
 
Monocular video depth estimation, which requires to estimate the depth of all  object instances appearing in given frames of a video that is captured using a single camera, has drawn more and more attention in recent past. 
Most of the existing approaches are developed for after-the-fact estimation, where the depth of images/frames which are to be estimated are accessible to the system (refer to Fig 1(a)). However, it is often required in many practical cases that the system can predict future depth estimation before the corresponding images/frames are observed (refer to Fig 1(b)). Future depth estimation (depth forecasting) is thus much more important than after-the-fact depth estimation in real-world applications like action interpretation \cite{nag2021few,nag2021temporal}, autonomous driving \cite{8803299}, etc.
For example, as illustrated in fig 1(e) due to the sudden right turn of the vehicle, a collison can be avoided if the depth of the car and free space around it is known to the driving system beforehand to perform an escape maneuver.

Video Depth forecasting is a challenging problem mostly due to uncertainty in appearance caused by object movement, occlusion and viewpoint.  
Qi et al. \cite{qi20193d} was the first to predict depth estimates for future time instants by reasoning over optical-flow (2D/3D), but assume access to additional information - depth images and semantic maps for past frames. More specifically, \cite{qi20193d} designed a complicated two-stage (prediction then refinement) architecture to jointly predict future motion,
RGB frames, depth maps, and semantic maps.
Very recently, Hu et al. \cite{hu2020probabilistic} adopted a conditional variational approach to model the inherent stochasticity of future prediction. The architecture learns rich representations that incorporate both local and global spatio-temporal context which are then used to forecast segmentation , depth and  flow.  While these approaches have attempted to reason the future using depth forecasting, both of them have some shortcomings : 
(a) both the approaches \cite{qi20193d,hu2020probabilistic} require labeled data like semantic, depth maps, motion estimates of the past/all frames during training/pre-training stage hence not generalizable to any video;
(b) although \cite{qi20193d} handles rigid scene assumptions for a very short-term (next frame), both \cite{qi20193d,hu2020probabilistic} fails to handle the scene consistency beyond that - resulting in artefacts of moving object.

\begin{figure*}[]
\vspace{-.05in}
\includegraphics[scale=0.13]{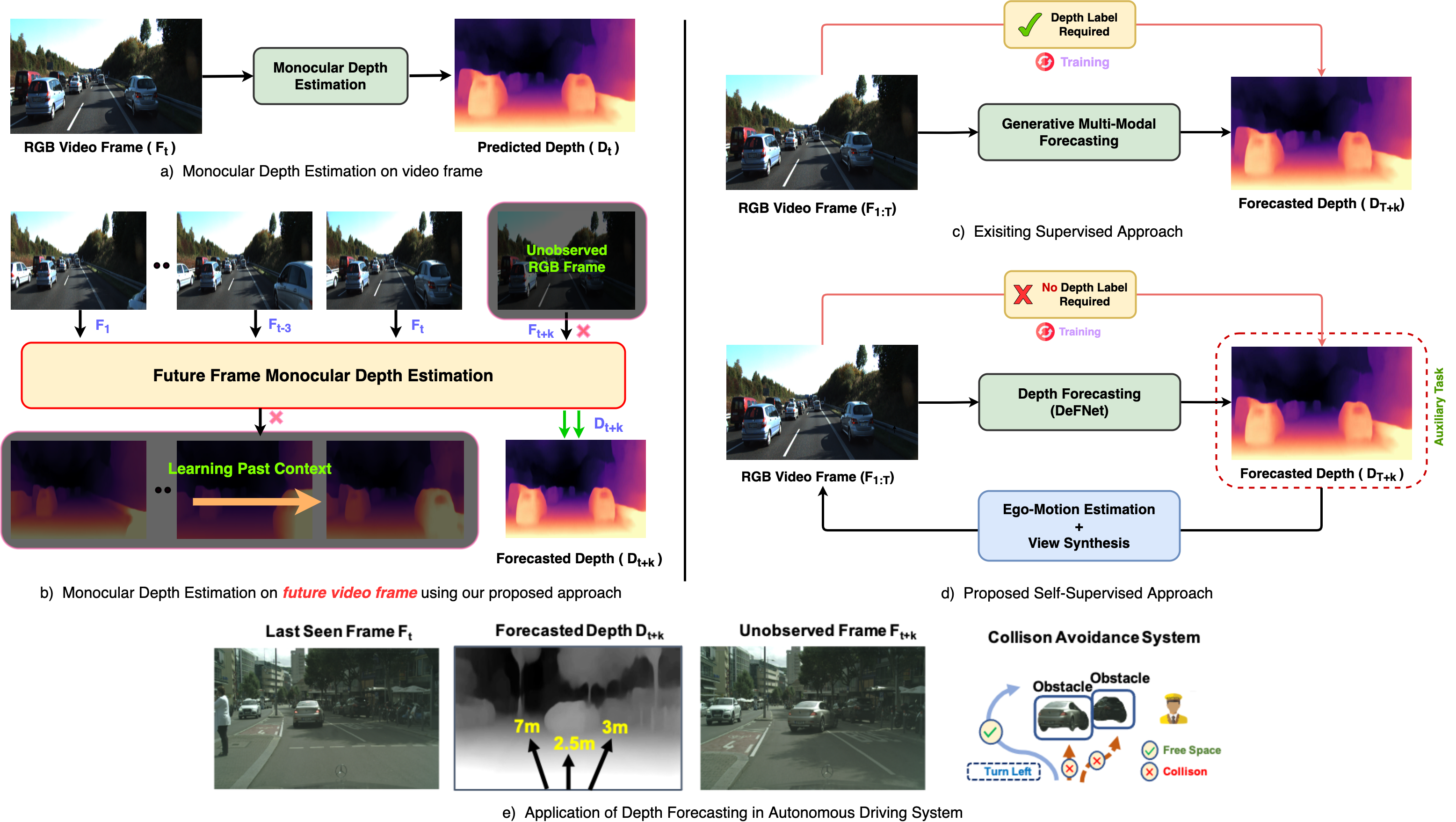}
\caption{\textbf{Overview of the problem.} Illustration of the standard problem of depth forecasting with existing baselines \cite{hu2020probabilistic} and its practical application in autonomous driving solution. } 
\label{fig:network}
\end{figure*}
In recent years, self-supervised methods have attracted increasing interests  in monocular depth estimation \cite{godard2019digging,godard2017unsupervised,luo2019every,zhou2017unsupervised} tasks thus making it a suitable alternative to supervised approaches \cite{cheng2020s,fu2018deep,lee2019big}.  In the absence of ground truth, one can still recover scene depth and ego-motion from
monocular video sequences using novel view synthesis \cite{godard2017unsupervised,garg2016unsupervised} (by jointly optimizing the depth and pose network). 
However, all of these self-supervised approaches are designed for after-the-fact depth estimation (fig 1(a)). 
Taking motivation from this, we propose a new self-supervised depth forecasting setting which is achieved by designing a novel Depth Forecasting Network and a new pipeline for Pose Estimation conditioned on the unseen future.
The Depth Forecasting Network (DeFNet) consists of a novel ConvGRU based feature forecasting block and a convLSTM based flow forecasting block. The forecasting blocks take aggregated features till time step $\{t\}$ and forecast a feature at time step $\{t+k\}$ which is then post-processed to form forecasted depth. Additionally, we also designed a new channel-attention based Pose Network for the forecasting task that enforces the network to leverage the pose w.r.t the unseen target frame at time $\{t+k\}$.
As a result of these changes, we modify the existing self-supervised after-the-fact setting to propose a new \textit{self-supervised forecasting setting}. Thus, we a) solve depth forecasting as an auxiliary task thereby eliminated the need of expensive depth labels hence making our approach generalizable to any form of video scene and b) handle the rigid scene-assumption for both short and mid-term forecast (by varying k in $\{t+k\}$).

\noindent \textbf{Contributions} In summary, our contributions are three-folds : \textbf{(1)} We proposed a novel multi-scale feature-level forecasting network (DeFNet) for depth forecasting \textbf{(2)} We also designed a new pose estimation network using Channel attention (PCAB) and adapted it for the forecasting task. \textbf{(3)} We formulated a new self-supervised setting for depth forecasting task that eliminates the need of depth annotation making it generalizable and scalable. Extensive experiments show that the proposed method yields new state-of-the-art performance on two benchmark datasets ( KITTI and Cityscapes ).

\section{Related Works}

\textbf{Self-Supervised Depth Estimation}:
A more promising substitute for supervised depth models \cite{casser2019depth, karsch2014depth, ranftl2016dense, wang2019web, bloesch2018codeslam, zhou2018deeptam,lee2019big, cheng2020s} is the self-supervised approach. A less constrained form of self-supervision is to use monocular videos, where consecutive temporal frames provide the training signal. 
In one of the first monocular self-supervised approaches, \cite{zhou2017unsupervised} trains a depth estimation network along with a separate pose network. Inspired by \cite{byravan2017se3}, \cite{vijayanarasimhan2017sfm} proposes a more sophisticated motion model using multiple motion masks. \cite{yin2018geonet} also decomposes motion into rigid and non-rigid components, using depth and optical flow to explain object motion. In the context of optical flow estimation, \cite{janai2018unsupervised} shows that it helps to explicitly model occlusion.
 Additional data such as pre-computed instance segmentation masks were used by \cite{casser2019depth} to help deal with moving objects. In another work, \cite{godard2017unsupervised} shows that the inclusion of a local structure based appearance loss \cite{wang2004image} significantly improves depth estimation performance compared to simple pairwise pixel differences \cite{xie2016deep3d,garg2016unsupervised,zhou2017unsupervised}. Self-supervision was achieved by \cite{godard2019digging} by jointly training a depth and pose network using novel view synthesis. Nevertheless, all of these existing self-supervised approaches are formulated for after-the-fact setting depth estimation. In this paper, we for the first time develop a self-supervised monocular depth solution for our new forecasting setting. 
\\\\
\noindent\textbf{Video Forecasting}:  
These approaches are useful in generating a future frame by learning past frame correspondence. 
Ranzato et al. \cite{ranzato2014video} introduces the first baseline in this domain by presenting a generative approach for the future frame prediction. Few works followed this \cite{mathieu2015deep,luo2017unsupervised, kalchbrenner2017video}, which improves the learning of past context using Long Short Term Memory (LSTM). 
Different from predicting frames, Jin et al. \cite{jin2017video} directly predicts the semantic label of the future frame i.e semantic forecasting by observing the past frames. Several works \cite{nilsson2018semantic,luc2017predicting} thereby followed that fused optical flow to model semantic forecast for longer durations. Along similar motivation to our work, semantic forecasting was addressed by forecasting intermediate latent representations of past observed frames using CNN \cite{luc2018predicting,saric2020warp} and ConvLSTMs \cite{sun2019predicting}. Depth forecasting was first addressed by \cite{qi20193d} as a part of future frame synthesis that can only predict the depth of next time instant. Recently, Hu  et al. \cite{hu2020probabilistic} addressed this task using generative modelling of a probabilistic future. However, both of these approaches require expensive depth annotation during training or pre-training stage to generate the depth forecast. Our approach is thus the first work that attempts to solve the depth forecasting task in a completely self-supervised fashion without any label requirement.

\section{Proposed Methodology}

We introduce the novel Depth Forecasting Network and describe overall design in Sec 3.1. In Sec 3.2, we describe our new pipeline for Pose Estimation and discuss about adapting it for the new forecasting setting, achieving self-supervision using novel-view synthesis. Finally, we discuss the learning objective as well as the inference process in Sec 3.3. The overall architecture is illustrated in Fig 2. 

\noindent \textbf{Problem Formulation:} 
Depth forecasting is defined as the task when given a video  $\textbf{V}$ with $\textit{t}$ frames $\{ I_{1},I_{2},...,I_{t}\} \in V$ and the corresponding optical flow $\textbf{M} \in \{ O_{1},O_{2},...,O_{t}\}$, we anticipate the depth $\textbf{D}_{t+k}$ at a fixed number of timesteps $\textit{k}$ from the last observed frame at time t. This anticipated depth corresponds to the depth of an unobserved future frame ${I}_{t+k}$. In this paper, we will discuss models for both short-term $(k = 5)$ and mid-term $(k = 10)$ prediction. For training, we sample 4 frames from $V$ and $M$ with frame interval 3 per batch : $V_{1:t} = \{I_{t-9},I_{t-6},I_{t-3},I_{t}\}$; $M_{1:t} = \{O_{t-9},O_{t-6},O_{t-3},O_{t}\}$ and feed it to our Depth Forecasting Network (DeFNet). Additionally, we pass $\{I_{t},I_{t+k}, I_{t+2k}\}$ as input to the Pose Estimation network. 

\subsection{Depth Forecasting Network}
Depth forecasting network primarily consists of (a) Feature encoder - to encode pyramidal features , (b) Forecasting module - to forecast the features, (c) Decoder module - to decode forecasted depth from the features and (d) Fusion block - which refines the depth by fusing flow forecast.

\noindent \textbf{Feature Encoder:} Our approach begins with a encoder each for RGB and Flow input that extracts the multi-scale pyramidal features. 
In general any encoder backbone can be used, but due to performance gain we chose a pre-trained ResNext \cite{kone2018hierarchical} as RGB Encoder backbone and a pretrained LiteFlowNet2.0 \cite{hui2019lightweight} as Flow Encoder backbone. Formally, given a video $V_{1:t}$ and optical flow $M_{1:t}$ having frames of size $W \times H$, the encoder extracts $\{F_{rgb}^{t-9}, F_{rgb}^{t-6},..,F_{rgb}^{t}\}$ for rgb and $\{F_{flow}^{t-9}, F_{flow}^{t-6},..,F_{flow}^{t}\}$ for flow image where $F^{t}$ indicates the pyramid features $\{P^{i}_{t}\}_{i=1}^L$ (L scales in total) extracted from t-th frame. $P^{l}_{t}$
is the feature of the l-th pyramid level at t-th frame having dimension 
$[\frac{W}{2^{l}},\frac{H}{2^{l}}]$. The resultant rgb and flow features are then aggregated along the temporal dimension to obtain $\textbf{F}^{1:t}_{rgb}$ and $\textbf{F}^{1:t}_{flow}$ as follows :
\vspace{-0.1in}
\begin{align}
\textbf{F}^{1:t}_{rgb} = F_{rgb}^{t-9} \oplus  F_{rgb}^{t-6}\oplus ..\oplus F_{rgb}^{t} ,
\hspace{0.2in} \textbf{F}^{1:t}_{flow} = F_{flow}^{t-9} \oplus F_{flow}^{t-6}\oplus..\oplus F_{flow}^{t}
\end{align}
where $\oplus$ indicates channel-wise concatenation of features. The resultant multi-scale feature maps ($\textbf{F}^{1:t}_{rgb}$ and $\textbf{F}^{1:t}_{flow}$) are simultaneously fed into the feature forecasting module to predict the features for time-step t+k at multiple-scales.

\begin{figure*}[!hbtp]
\begin{center}
  \includegraphics[scale=0.21]{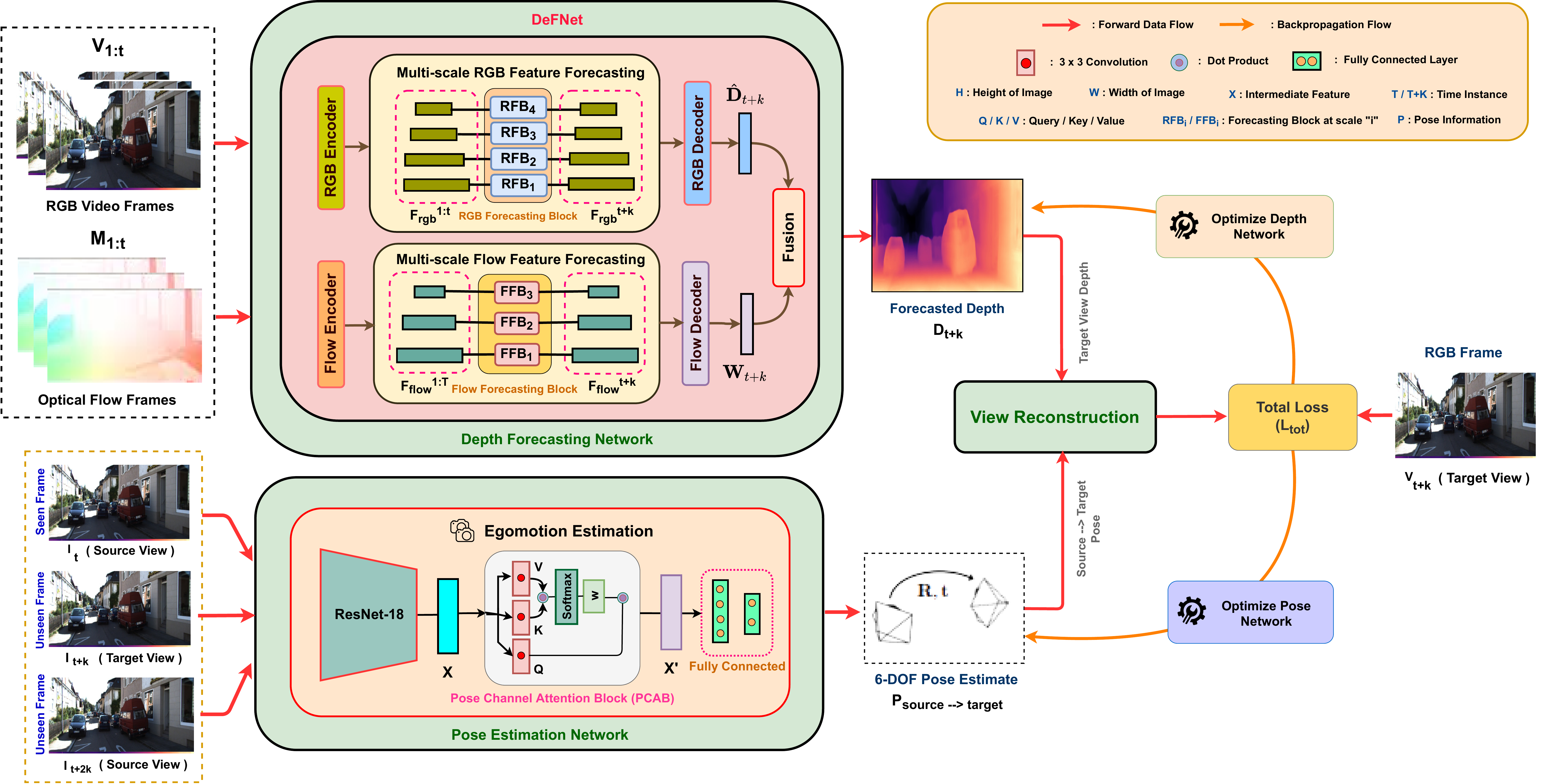}
\end{center}
\vspace{-.20in}
\caption{\textbf{Overview
of our proposed self-supervised Depth Forecasting approach}. Our novel Depth Forecasting network (DeFNet) takes in rgb and flow frames and forecasts depth $D_{t+k}$. The Pose Estimation Network takes in source and target input frames to produce pose estimate P conditioned on target frame unobserved by the depth network. The View reconstruction module reconstructs back the target view using depth $D_{t+k}$ and pose P. The reconstruction cost obtained from the target view and reconstructed target view acts as the supervision signal to train both the depth and pose networks.}
\label{fig:archi}
\end{figure*}

\noindent \textbf{Feature Forecasting Module:}
Our feature forecasting module receives processed input
rgb and flow features ($\textbf{F}^{1:t}_{rgb} / \textbf{F}^{1:t}_{flow}$) and directly regresses the future features $\textbf{F}^{t+k}_{rgb}, \textbf{F}^{t+k}_{flow}$ respectively. More specifically, the forecasting module learns a mapping $\phi$ between the pyramid
features extracted from past video frames and future frames.  
This mapping distinguishes our forecasting setting from the standard after-the-fact setting by predicting features for unobserved frames. The features predicted for the unobserved frame $I_{t+k}$ is formulated as :

\begin{align}
    \textbf{F}^{t+k}_{s} = \phi_{s}(\textbf{F}^{1:t}_{s}) \hspace{0.1in}where \hspace{0.1in} s \in \{rgb,flow\}
\end{align}

The rgb forecasting function $\phi_{rgb}$ consists of a feature forecasting block (termed as RFB) at each pyramidal level as illustrated in Fig 3. Each RFB block comprises of a series of convGRUs (cGRU) \cite{ballas2015delving} to model spatio-temporal relation (intra-level) among the features of the same pyramid level. In addition, our method also introduces connections (inter-level) between different cGRUs (in different RFBs) in order to capture the spatio-temporal context across different levels. The inter-connected cGRUs form our model (i.e., the mapping $\phi_{rgb}$ in formula (2)) for predicting future rgb features.
More details on intra-level and inter-level cGRU is provided in appendix. Similar to rgb forecasting, the flow forecasting function $\phi_{flow}$ consists of a flow feature forecasting block (termed as FFB) at each pyramidal level. Each FFB block consists of a single convLSTM \cite{terwilliger2019recurrent} and a 1-D convolution. For simplicity and ease of training, we do not consider inter or intra-level connections among different FFBs at each level. The extracted flow features $\textbf{F}^{1:t}_{flow}$ are passed as 
input to the FFB with a $3 \times 3$ kernel and $1 \times 1$ padding.
It then produces the flow forecasted feature $\textbf{F}^{t+k}_{flow}$ using a
single $1 \times 1$ convolution layer after the ConvLSTM to reduce the channels for the flow field. 

\noindent \textbf{Feature Decoder:}
The input of our feature decoder is the predictive features ($\textbf{F}^{t+k}_{rgb}/\textbf{F}^{t+k}_{flow}$)
forecasted by our feature-forecasting module. As illustrated in fig 3, the  decoders progressively predicts depth and flow features at each pyramidal level in a coarse-to-fine manner. The rgb decoder first predicts the low resolution depth map $\hat{\textbf{D}}^L_{t+k}$ at the top-level with size $\frac{W}{2^L} \times \frac{H}{2^L}$ as the initial scene layout using a convolution operation.  
As illustrated in fig 3(b), the pyramidal feature $\textbf{P}^{i}_{t+k}$ and the depth map $\hat{\textbf{D}}^{i+1}_{t+k}$ predicted at the upper level are integrated together to produce a refined depth map $\hat{\textbf{D}}^{i}_{t+k}$ in the current scale. 
The output depth representation $\hat{\textbf{D}}_{t+k}$ thereby contains the refined depth from multiple scales that carry both high-level scene information and low-level detail information. Similarly, for the flow decoder, we follow the cascaded flow refinement design of \cite{hui2019lightweight} at each pyramid level. Given the forecasted features $\textbf{F}^{t+k}_{flow}$, it obtains refined flow $\textbf{W}_{t+k}$ using a 2-stage refinement strategy. First, the pixel by-pixel matching of high-level feature vectors yields a coarse flow estimate. This is achieved by upsampling the previous pyramid level flow $\textbf{W}^{i+1}_{t+k}$ and warping it on to the current level flow feature $P_{t+k}^i$.
Second, a subsequent refinement on the coarse flow further improves it to a refined flow. A correlation between the warped coarse flow and the current level feature $P_{t+k}^{i}$ results in a refined $\textbf{W}^{i}_{t+k}$. The final decoded depth $\hat{\textbf{D}}_{t+k}$ and flow $\textbf{W}_{t+k}$ is obtained from last (bottom) pyramidal layer output.
\begin{figure*}[!hbtp]
\begin{center}
  \includegraphics[scale=0.17]{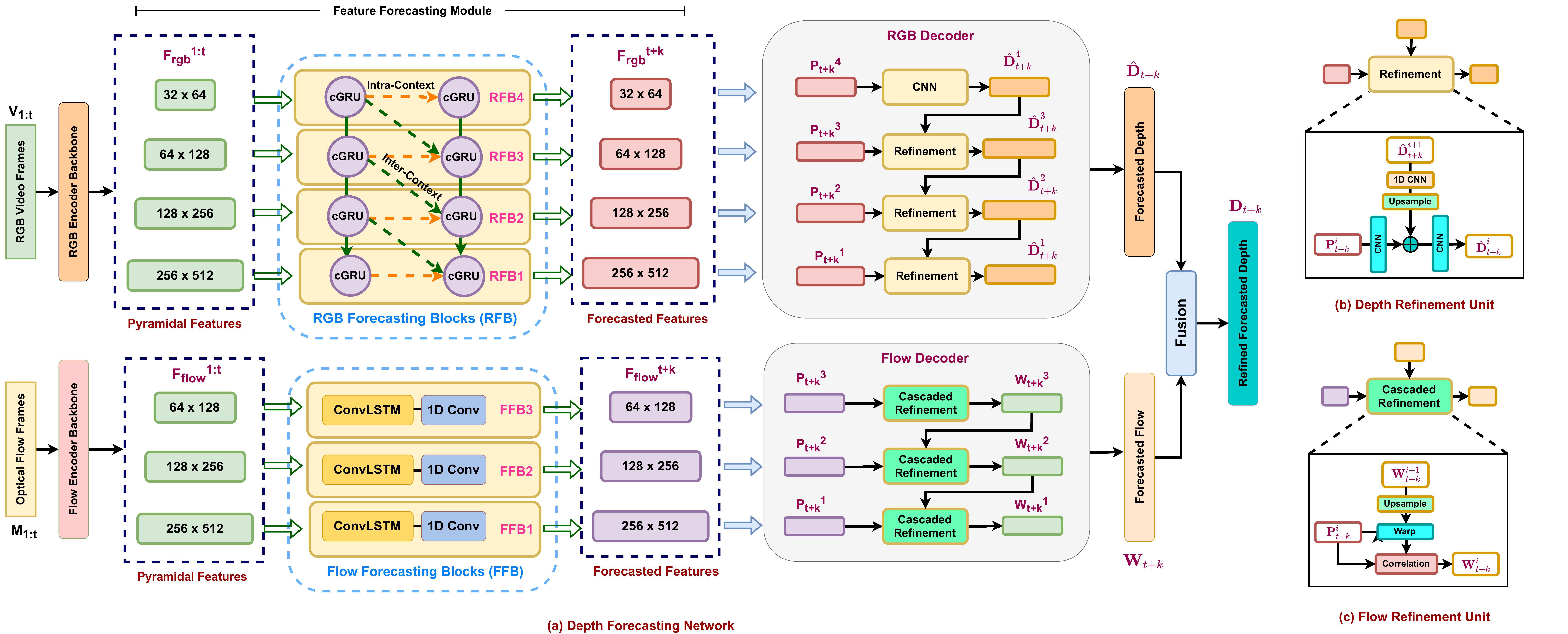}
\end{center}
\vspace{-.20in}
\caption{\textbf{Illustration
of our novel Depth Forecasting Network (DeFNet)}It consists of a encoder , a forecasting and decoder module. The encoded pyramidal features are forecasted using the feature forecasting blocks and then decoded in a coarse-to-fine manner. The refined forecasted depth is obtained by fusing both rgb and flow forecast. The refinement units of depth and flow decoders are illustrated in (b) and (c) respectively.}
\vspace{-.15in}
\label{fig:archi}
\end{figure*}

\noindent \textbf{Fusion:} Recall, that the forecasted depth estimate $\hat{\textbf{D}}_{t+k}$ already carries structural geometries and the forecasted flow information $\textbf{W}_{t+k}$ carries the object motions for the unobserved scene. However, the occluded objects (not observed in the past frames) tend to lose its structure in the forecasted depth $\hat{\textbf{D}}_{t+k}$ and thus can suffer from incomplete depth estimates and self-occlusions. Since flow information can still predict the motion of such occluded pixels , fusing both forecasted depth and flow helps our network to make meaningful assumptions of the future scene by considering observed motion of depth feature activations. The final forecasted depth $\textbf{D}_{t+k}$ is obtained by using a single learnable convolution \cite{henriques2017warped} layer $\Theta$ as follows : 
\vspace{-0.10in}
\begin{align}
    \textbf{D}_{t+k} = \sigma(\Theta(\left \|  \hat{\textbf{D}}_{t+k} \right \|, \left \| \textbf{W}_{t+k} \right \|))
\end{align}
where $\sigma$ is sigmoid operation. It is to be noted that both $\hat{\textbf{D}}_{t+k},\textbf{W}_{t+k}$ are normalized before passing into $\Theta$.  For non-translational motion, $\Theta$ can compensate for some geometric distortions by learning some depth for the occluded pixels thus producing a refined forecast.

\subsection{View Reconstruction using Future Conditioned Ego-Motion}
The key supervision signal for our network comes from the task of \textit{novel view synthesis}: given one input view of a scene, synthesize a new image of the scene seen from a different camera pose. We can synthesize a target view given a per-pixel depth in that image, plus the pose and visibility in a nearby view. 
As we will show next, this view-synthesis approach is re-designed for our forecasting setting to achieve self-supervision. 

 \noindent \textbf{Ego-Motion:} The ego-motion network shown in the bottom half of fig. 2 estimates relative pose $P \in \mathbb{SE}(3)$ introduced by motion fields across frames. Specifically, it learns a function $\Phi$ which is a CNN for predicting the camera motion of input frames. For instance, given a sequence of images $(I_{t-1},I_{t},I_{t+1})$ with target frame $I_{t}$ and temporally neighboring reference frames $I_{t-1},I_{t+1}$, $\Phi$ predicts a vector $\textbf{P}_{t-1 \rightarrow t}$ and $\textbf{P}_{t+1\rightarrow t}$ comprising of translation ($t_{x}, t_{y}, t_{z}$) and rotation ($\varrho , \theta, \psi$) parameters of the camera between the frames. Mathematically, it is denoted as 
 \begin{align}
     \textbf{P}_{so \rightarrow ta} = \Phi(I_{so},I_{ta}) \hspace{0.1in}where \hspace{0.1in} I_{so} \in \{I_{t},I_{t+1}\} , I_{ta} \in \{I_{t}\}
 \end{align}
 where $I_{so}$ denotes source images and $I_{ta}$ denotes target images. 
 This formulation is commonly used in after-the-fact setting for pose estimation where pose is calculated for the frames seen by the depth network. In order to adapt this for our forecasting setting we used an interesting trick : instead of passing a sequence of consecutive images $(I_{t-1},I_{t},I_{t+1})$ , we input $(I_{t},I_{t+k},I_{t+2k})$ (i.e $I_{so} \in \{I_{t},I_{t+2k}\}$ and $I_{ta} \in \{I_{t+k}\}$ in formula 4 ). It is interesting to note that the frames $I_{t+k},I_{t+2k}$ are unobserved video frames for the depth network. We argue that by passing these unobserved frames, the Pose-Network $\Phi$ predicts  $\textbf{P}_{t \rightarrow t+k}$ and $\textbf{P}_{t+2k\rightarrow t+k}$ which enforces the network to learn the camera translation ($t_{x}, t_{y}, t_{z}$) and rotation ($\varrho , \theta, \psi$) parameters conditioned on the target frame $I_{t+k}$. This leverages a pose information that captures both seen pose context ($\textbf{P}_{t \rightarrow t+k}$) and also unseen pose context ($\textbf{P}_{t+2k \rightarrow t+k}$) while making an estimate.
 
 \noindent \textbf{Network Design:} The pose-network $\Phi$ consists of a ResNet-18 \cite{he2016deep} encoder , a channel attention block ( referred as PCAB) followed by 2 fully connected layers to predict $6-DOF$ output. Given a sequence of images $\{I_{t}, I_{t+k}, I_{t+2k}\}$ the encoder takes inputs in pairs and outputs a feature $X \in \mathbb{R}^{7 \times 7 \times 512}$. Since we are not passing consecutive frames ($k \geqslant 1$ in $I_{t+k},I_{t+2k}$) through the network it is prone to uncertainties introduced by moving objects, occlusions. 
 Such error prone regions between the target and source image feature is highlighed using our channel attention PCAB.
 The PCAB primarily consists of 3 convolution layers having kernel size 3 to obtain Q/K/V which is used to obtain the channel weight vector $\omega$. The weighted pose feature $X^{'}$ is obtained by multiplying weight vector $\omega$ with the input feature $X$. More details on PCAB is provided in appendix. The weighted feature $X^{'}$ is then passed into the fully connected layers to obtain 6-DOF pose $\textbf{P}$. 

 \noindent \textbf{View Reconstruction:} Given a target and source image pair ($I_{ta}, I_{so}$) with known transformation ($\textbf{P}_{so \rightarrow ta}$) between the images and the forecasted depth map ($\textbf{D}_{ta}$), the target image can be reconstructed by sampling pixels from the source image through image inverse warping.
Let $j_{ta}$ denotes the 2D homogeneous coordinate of an pixel in frame $I_{ta}$ and K denotes the intrinsic camera matrix. We can compute the corresponding point of $j_{ta}$ (denoted as $j_{so}$) in frame $I_{so}$ using the following equation: 
 \begin{align}
    j_{so} \sim KP_{so \rightarrow ta}D_{ta}(j_{ta})K^{-1}j_{ta}
 \end{align}
The reconstructed target image $I_{so \rightarrow ta}$ is obtained by populating the value of $I_{ta}(j_{so})$ using differentiable inverse warping \cite{jaderberg2015spatial}. This reconstruction is feasible for monocular videos which is mainly based on the assumptions that the scene is static without moving object, the vision difference is caused by the camera pose change and no new object appears into the view between the target view and the source views. But this is hard to satisfy for all training sequences collected in real world. Inspired by \cite{godard2019digging,cheng2020s}, we use a auto-mask (denoted by $A$) that filters out such pixels which do not change appearance from one frame to the next in the sequence. 
It is described as follows : 
\begin{align}
    A_{so \rightarrow ta} = \underset{so}{min}L_{pe}(I_{ta},I_{so \rightarrow ta}) < \underset{so}{min}L_{pe}(I_{ta},I_{so})
\end{align}
where $L_{pe}$ denotes photometric reprojection loss (described in Section 3.3). This mask $\textbf{A}$ forces the pose-network to ignore objects which move at the same velocity as the camera, and even to ignore whole frames in monocular videos when the camera stops moving. 

\subsection{Learning Objective and Inference :} 
We train our network (both depth and pose) using the reconstruction error between the reconstructed target $I_{so \rightarrow ta}$ and original target $I_{ta}$. The loss components are defined as follows :
\\
(a) \textbf{Masked Photometric($L_{mpe}$) }: Photometric loss is a standard L1 loss calculated between the reconstructed view $I_{so \rightarrow ta}$ and target view $I_{ta}$ denoted as $L_{pe}$. We formulate the mask photometric loss $L_{mpe}$ by performing the dot product with mask $L_{pe}$ as follows: 
\begin{equation}
 L_{mpe} = \underset{s}{\sum}(A_{s}.L_{pe}) , s \in(so \rightarrow ta, ta \rightarrow so) 
\vspace{-0.1in}
\end{equation} 
 
 \noindent(b) \textbf{Dissimilarity ($L_{ds}$):} It calculates the dissimilarity between source and target views by also being differentiable. It is denoted by 
 \begin{equation}
    L_{ds} = 0.5*(1-SSIM(I_{so \rightarrow ta},I_{ta}))
 \end{equation}
 
 \noindent(c) \textbf{Smoothness ($L_{sm})$:} It enforces DeFNet to produce sharp edge distribution while producing smoother depth. It is denoted by $L_{sm}$ and has similar formulation as in \cite{garg2016unsupervised}. 
 \\
 (d) \textbf{Pose Consistency ($L_{pc}$):}It a simple L1 loss that ensures the 'past' and 'future' inter-frame translations are consistent with each other. It is represented as 
 \begin{equation}
     L_{pc} = \left \| P_{t,t+k} - P_{t+2k,t+k}\right \|_{1}
 \end{equation}. 
 We combine all the losses to form our final training objective as follows : $L_{tot} = \alpha.L_{mpe} + (1-\alpha).L_{ds} + \lambda.L_{sm} + \gamma.L_{pc}$ where $\alpha , \lambda , \gamma$ are hyperparameters. Following standard practise in self-supervised after-the-fact setting \cite{godard2019digging}, both the DeFNet and Pose networks are jointly optimized using this loss. More details on loss are given in the appendix.

During Inference, we discard the pose estimation network similar to after-the-fact setting \cite{godard2019digging} and only use the depth forecasting network DeFNet to estimate the forecasted depth of testing videos.

\section{Experiments}
\noindent \textbf{Dataset:} We use the \textbf{KITTI} \cite{geiger2013vision} and \textbf{Cityscapes} \cite{cordts2016cityscapes} datasets to evaluate our method for outdoor scenes. For KITTI, we adopt the training protocols used in Eigen et al. \cite{eigen2014depth}, and specifically, we use the KITTI Eigen splits that contain 22600 train, 888 validation, and 697 test stereo image pairs. For Cityscapes, it is a more challenging dataset with many dynamic scenes. With a few exceptions \cite{pilzer2018unsupervised,casser2019unsupervised,hu2020probabilistic} it has not been used
for depth estimation evaluation. It has 38675 training examples. We use depth from the disparity data for evaluation on a standard evaluation set of 1250 samples \cite{pilzer2018unsupervised,casser2019unsupervised}. We evaluate on both these datasets following the same evaluation strategy used in semantic forecasting \cite{luc2017predicting,luc2018predicting}.

\noindent \textbf{Implementation Details:}  We implement our framework in PyTorch and conduct experiments on a 11 GB Nvidia RTX 2080-Ti GPU. The ResNext \cite{xie2017aggregated} backbone is pre-trained on ImageNet \cite{deng2009imagenet} using batch normalisation.  The input dimensions of
the video frames are set as $512 \times 1024$ i.e $H^{'}= 512$ and $W^{'}= 1024$. Optical Flow for the video frames are computed using the flow network
in FlowNet2.0 \cite{ilg2017flownet}. Our DeFNet and Ego-Motion Network is jointly trained for 30 epochs with learning rate of $0.0001$ using Adam Optimizer \cite{kingma2014adam}. The entire network takes $53$ mins to converge. The learning objective hyperparameters $\alpha / \lambda / \gamma$ are set to 0.4/0.5/0.6. For fair comparison, we follow existing work \cite{hu2020probabilistic}, where we consider k = 5 and k = 10 frames in the future (corresponding to respectively 0.29s and 0.59s). The mean inference speed of our DeFNet on image of size $512 \times 1024$ is $13$ FPS on 2080GTX GPU.

\noindent \textbf{Evaluation Metric:} We quantify our approach following standard monocular depth estimation setting in short and mid term setting. Specifically, we report the square relative error ( Sq Rel), absolute relative error (Abs Rel), root mean square error (RMSE), log scale invariant RMSE (RMSE-Log) and the amount of inliers ($\delta$) for both KITTI and Cityscapes dataset. We refer the reader to \cite{eigen2014depth} or appendix for a complete description of these metrics. 
\vspace{-0.10in}
\subsection{Comparison with the state-of-the-art methods}
\noindent \textbf{Competitors:} For comparative comparison, we consider a multi-scale instance segmentation model \cite{sun2019predicting}, a unsupervised depth estimation model \cite{godard2019digging} and the latest depth forecasting approaches \cite{hu2020probabilistic,qi20193d}. Because \cite{godard2019digging} is not designed for forecasting task, we add a basic F2F\cite{luc2018predicting} before the decoder of the depth network. Since KITTI does not have vehicle control data, we cannot train \cite{hu2020probabilistic} on KITTI, instead we compare another similar approach \cite{qi20193d} for both short and mid-term forecast. Following \cite{luc2018predicting,hu2020probabilistic}, we report the accuracy of oracle as a upper bound and we use a trivial copy baseline as a lower bound.

\begin{table}[]
\resizebox{1\columnwidth}{!}{
\begin{tabular}{ccccccccc}
\hline
\multicolumn{9}{c}{\textbf{KITTI}}                                                                                                             \\ \hline
\multicolumn{1}{c|}{\multirow{2}{*}{Method}} & \multicolumn{1}{c|}{\multirow{2}{*}{Supervision}} & \multicolumn{3}{c|}{Higher is better}                                                                                                                                   & \multicolumn{4}{c}{Lower is better}                                                              \\ \cline{3-9} 
\multicolumn{1}{c|}{}                        & \multicolumn{1}{c|}{}                             & \multicolumn{1}{c|}{$\delta$ < 1.25} & \multicolumn{1}{c|}{$\delta$  < $1.25^2$} & \multicolumn{1}{c|}{$\delta$  < $1.25^3$} & \multicolumn{1}{c|}{Abs Rel} & \multicolumn{1}{c|}{Sq Rel} & \multicolumn{1}{c|}{RMSE} & RMSE-Log \\ \hline
\multicolumn{1}{c|}{Oracle}                  & \multicolumn{1}{c|}{-}                            & \multicolumn{1}{c|}{0.888}                   & \multicolumn{1}{c|}{0.970}                                      & \multicolumn{1}{c|}{0.984}                                      & \multicolumn{1}{c|}{0.097}      & \multicolumn{1}{c|}{0.734}     & \multicolumn{1}{c|}{4.442}   & 0.187       \\
\multicolumn{1}{c|}{Copy Last}               & \multicolumn{1}{c|}{-}                            & \multicolumn{1}{c|}{0.816}                   & \multicolumn{1}{c|}{0.941}                                      & \multicolumn{1}{c|}{0.976}                                      & \multicolumn{1}{c|}{0.141}      & \multicolumn{1}{c|}{1.029}     & \multicolumn{1}{c|}{5.350}   & 0.216       \\ \hline
\multicolumn{1}{c|}{\text{Qi et al.} \cite{qi20193d}}                & \multicolumn{1}{c|}{Supervised}                   & \multicolumn{1}{c|}{-}                   & \multicolumn{1}{c|}{-}                                      & \multicolumn{1}{c|}{-}                                      & \multicolumn{1}{c|}{0.108}      & \multicolumn{1}{c|}{0.806}     & \multicolumn{1}{c|}{4.630}   & 0.193       \\
\multicolumn{1}{c|}{Sun et al. \cite{sun2019predicting}}              & \multicolumn{1}{c|}{Supervised}                   & \multicolumn{1}{c|}{0.852}                   & \multicolumn{1}{c|}{0.947}                                      & \multicolumn{1}{c|}{0.977}                                      & \multicolumn{1}{c|}{0.112}      & \multicolumn{1}{c|}{0.875}     & \multicolumn{1}{c|}{4.958}   & 0.207       \\
\multicolumn{1}{c|}{Goddard et al. \cite{godard2019digging}}              & \multicolumn{1}{c|}{Unsupervised}                 & \multicolumn{1}{c|}{0.877}                   & \multicolumn{1}{c|}{0.959}                                      & \multicolumn{1}{c|}{0.981}                                      & \multicolumn{1}{c|}{0.115}      & \multicolumn{1}{c|}{0.903}     & \multicolumn{1}{c|}{4.863}   & 0.193       \\ \hline
\multicolumn{1}{c|}{Ours}                    & \multicolumn{1}{c|}{Unsupervised}                 & \multicolumn{1}{c|}{\textbf{0.878}}                   & \multicolumn{1}{c|}{0.953}                                      & \multicolumn{1}{c|}{\textbf{0.983}}                                      & \multicolumn{1}{c|}{\textbf{0.107}}      & \multicolumn{1}{c|}{0.849}     & \multicolumn{1}{c|}{\textbf{4.614}}   & \textbf{0.182}       \\
\multicolumn{1}{c|}{Ours w/o Automask}                    & \multicolumn{1}{c|}{Unsupervised}                 & \multicolumn{1}{c|}{0.803 }                   & \multicolumn{1}{c|}{\textbf{0.960} }                                      & \multicolumn{1}{c|}{0.986}                                      & \multicolumn{1}{c|}{0.113 }      & \multicolumn{1}{c|}{\textbf{0.741} }     & \multicolumn{1}{c|}{4.621 }   & 0.189        \\
\multicolumn{1}{c|}{Ours w/o PCAB}                    & \multicolumn{1}{c|}{Unsupervised}                 & \multicolumn{1}{c|}{0.864 }                   & \multicolumn{1}{c|}{0.954 }                                      & \multicolumn{1}{c|}{0.979}                                      & \multicolumn{1}{c|}{0.111}      & \multicolumn{1}{c|}{ 0.867 }     & \multicolumn{1}{c|}{4.714 }   & 0.199        \\
\multicolumn{1}{c|}{Ours w/o PCAB + Automask}                    & \multicolumn{1}{c|}{Unsupervised}                 & \multicolumn{1}{c|}{0.844 }                   & \multicolumn{1}{c|}{0.941}                                      & \multicolumn{1}{c|}{0.978}                                      & \multicolumn{1}{c|}{0.119 }      & \multicolumn{1}{c|}{1.201 }     & \multicolumn{1}{c|}{5.888 }   & 0.208        \\
\multicolumn{1}{c|}{Ours w/o Flow Forecast}                    & \multicolumn{1}{c|}{Unsupervised}                 & \multicolumn{1}{c|}{0.859 }                   & \multicolumn{1}{c|}{0.947}                                      & \multicolumn{1}{c|}{0.980}                                      & \multicolumn{1}{c|}{0.113 }      & \multicolumn{1}{c|}{0.894 }     & \multicolumn{1}{c|}{4.804 }   & 0.208        \\
\hline
\multicolumn{9}{c}{\textbf{Cityscapes}}                                                                       \\ \hline
\multicolumn{1}{c|}{\multirow{2}{*}{Method}} & \multicolumn{1}{c|}{\multirow{2}{*}{Supervision}} & \multicolumn{3}{c|}{Higher is better}                                                                                                                                   & \multicolumn{4}{c}{Lower is better}                                                              \\ \cline{3-9} 
\multicolumn{1}{c|}{}                        & \multicolumn{1}{c|}{}                             & \multicolumn{1}{c|}{$\delta$ < $1.25$} & \multicolumn{1}{c|}{$\delta$  <  $1.25^2$} & \multicolumn{1}{c|}{$\delta$  <  $1.25^3$} & \multicolumn{1}{c|}{Abs Rel} & \multicolumn{1}{c|}{Sq Rel} & \multicolumn{1}{c|}{RMSE} & RMSE-Log \\ \hline
\multicolumn{1}{c|}{Oracle}                  & \multicolumn{1}{c|}{-}                            & \multicolumn{1}{c|}{0.836 }                   & \multicolumn{1}{c|}{0.943 }                                      & \multicolumn{1}{c|}{0.974}                                      & \multicolumn{1}{c|}{0.127 }      & \multicolumn{1}{c|}{1.031 }     & \multicolumn{1}{c|}{5.266 }   & 0.221        \\
\multicolumn{1}{c|}{Copy Last}               & \multicolumn{1}{c|}{-}                            & \multicolumn{1}{c|}{0.765 }                   & \multicolumn{1}{c|}{0.893 }                                      & \multicolumn{1}{c|}{0.940}                                      & \multicolumn{1}{c|}{0.257 }      & \multicolumn{1}{c|}{4.238}     & \multicolumn{1}{c|}{7.273 }   & 0.448       \\ \hline
\multicolumn{1}{c|}{Qi et al. \cite{qi20193d}}                & \multicolumn{1}{c|}{Supervised}                   & \multicolumn{1}{c|}{0.678}                   & \multicolumn{1}{c|}{0.885}                                      & \multicolumn{1}{c|}{0.957}                                      & \multicolumn{1}{c|}{0.208}      & \multicolumn{1}{c|}{1.768}     & \multicolumn{1}{c|}{6.865}   & 0.283 
\\
\multicolumn{1}{c|}{Hu et al. \cite{hu2020probabilistic}}                & \multicolumn{1}{c|}{Supervised}                   & \multicolumn{1}{c|}{0.725}                   & \multicolumn{1}{c|}{0.906}                                      & \multicolumn{1}{c|}{0.963}                                      & \multicolumn{1}{c|}{0.182 }      & \multicolumn{1}{c|}{1.481}     & \multicolumn{1}{c|}{6.501 }   & 0.267        \\
\multicolumn{1}{c|}{Sun et al. \cite{sun2019predicting}}                & \multicolumn{1}{c|}{Supervised}                   & \multicolumn{1}{c|}{0.801}                   & \multicolumn{1}{c|}{0.913}                                      & \multicolumn{1}{c|}{0.950}                                      & \multicolumn{1}{c|}{0.227}      & \multicolumn{1}{c|}{3.80}     & \multicolumn{1}{c|}{6.91}   & 0.414  
\\
\multicolumn{1}{c|}{Goddard et al. \cite{godard2019digging}}              & \multicolumn{1}{c|}{Unsupervised}                 & \multicolumn{1}{c|}{\textbf{0.836}}                   & \multicolumn{1}{c|}{0.930}                                      & \multicolumn{1}{c|}{0.958}                                      & \multicolumn{1}{c|}{0.193}      & \multicolumn{1}{c|}{1.438}     & \multicolumn{1}{c|}{5.887}   & 0.234       \\ \hline
\multicolumn{1}{c|}{Ours}                    & \multicolumn{1}{c|}{Unsupervised}                 & \multicolumn{1}{c|}{0.793}                   & \multicolumn{1}{c|}{\textbf{0.931}}                                      & \multicolumn{1}{c|}{\textbf{0.973}}                                      & \multicolumn{1}{c|}{\textbf{0.174}}      & \multicolumn{1}{c|}{\textbf{1.296}}     & \multicolumn{1}{c|}{\textbf{5.857}}   & \textbf{0.233}       \\
\multicolumn{1}{c|}{Ours w/o Automask}                    & \multicolumn{1}{c|}{Unsupervised}                 & \multicolumn{1}{c|}{0.680 }                   & \multicolumn{1}{c|}{0.898}                                      & \multicolumn{1}{c|}{0.967}                                      & \multicolumn{1}{c|}{0.201 }      & \multicolumn{1}{c|}{1.584 }     & \multicolumn{1}{c|}{6.471 }   & 0.273        \\
\multicolumn{1}{c|}{Ours w/o PCAB}                    & \multicolumn{1}{c|}{Unsupervised}                 & \multicolumn{1}{c|}{0.784}                   & \multicolumn{1}{c|}{0.916}                                      & \multicolumn{1}{c|}{0.961}                                      & \multicolumn{1}{c|}{0.198}      & \multicolumn{1}{c|}{1.438}     & \multicolumn{1}{c|}{6.216}   & 0.270       \\
\multicolumn{1}{c|}{Ours w/o PCAB + Automask}                    & \multicolumn{1}{c|}{Unsupervised}                 & \multicolumn{1}{c|}{0.776 }                   & \multicolumn{1}{c|}{0.903 }                                      & \multicolumn{1}{c|}{0.949}                                      & \multicolumn{1}{c|}{0.234 }      & \multicolumn{1}{c|}{3.776}     & \multicolumn{1}{c|}{7.104 }   & 0.416       \\
\multicolumn{1}{c|}{Ours w/o Flow Forecast}                    & \multicolumn{1}{c|}{Unsupervised}                 & \multicolumn{1}{c|}{0.730 }                   & \multicolumn{1}{c|}{0.919 }                                      & \multicolumn{1}{c|}{0.958}                                      & \multicolumn{1}{c|}{0.189 }      & \multicolumn{1}{c|}{1.533}     & \multicolumn{1}{c|}{6.315}   & 0.279       \\
\hline
\end{tabular}

}
\caption{\textbf{Quantitative Evaluation} of our DeFNet model with existing approaches and baselines on KITTI val (eigen split) and Cityscapes dataset for short-term forecasting.}
\vspace{-0.15in}
\end{table}

\noindent \textbf{Results:} The short-term forecast results are compared in Tab 1. Refer to appendix for mid-term forecast results. It is evident that our method outperforms unsupervised approaches alongside performing competitively with existing supervised approaches for both the forecasts. This suggests the superiority of our forecasting model design over existing approaches. 
As illustrated in fig 4., in short-term forecast on KITTI our method gains by $\sim 1-5\%$ in Abs Rel over a supervised methods \cite{qi20193d,hu2020probabilistic} which is a significant contribution while beating strongly customised unsupervised baseline by $\sim 5-8\%$. 
For mid-term forecasts, \cite{hu2020probabilistic,godard2019digging} performs competitively as they use convLSTMs however being second-best to ours in most of the metrics. Our method performs better in mid-term forecast probably due to the PCAB block and multi-scale features. Interestingly, the variant of Ours without PCAB and Automask (last row of Tab 1) performs worse than adapted baseline of \cite{godard2019digging}. This indicates that vanilla PoseCNN along with cGRU cannot boost the performance alone.  In cityscapes, we similarly observe that our method still gains over other approaches by $\sim 8\%$ while \cite{sun2019predicting} performs the worst among all. This is because of the challenging scenes in the dataset and inability to handle occlusion. On the other hand, \cite{hu2020probabilistic} performs better than \cite{qi20193d,godard2019digging} suggesting that probabilistic modeling of harder scenes is a solution. Our variant without the flow forecast shows a sharp decrease ($\sim 8\%/11\%$) in performance for short/mid-term forecasts suggesting that 2D motion is useful in refining the depth in occluded scenes particularly in mid-term forecast. Please refer appendix for more qualitative and quantitative evaluation. 

\begin{figure*}[!hbt]
\begin{center}
  \includegraphics[scale=0.26]{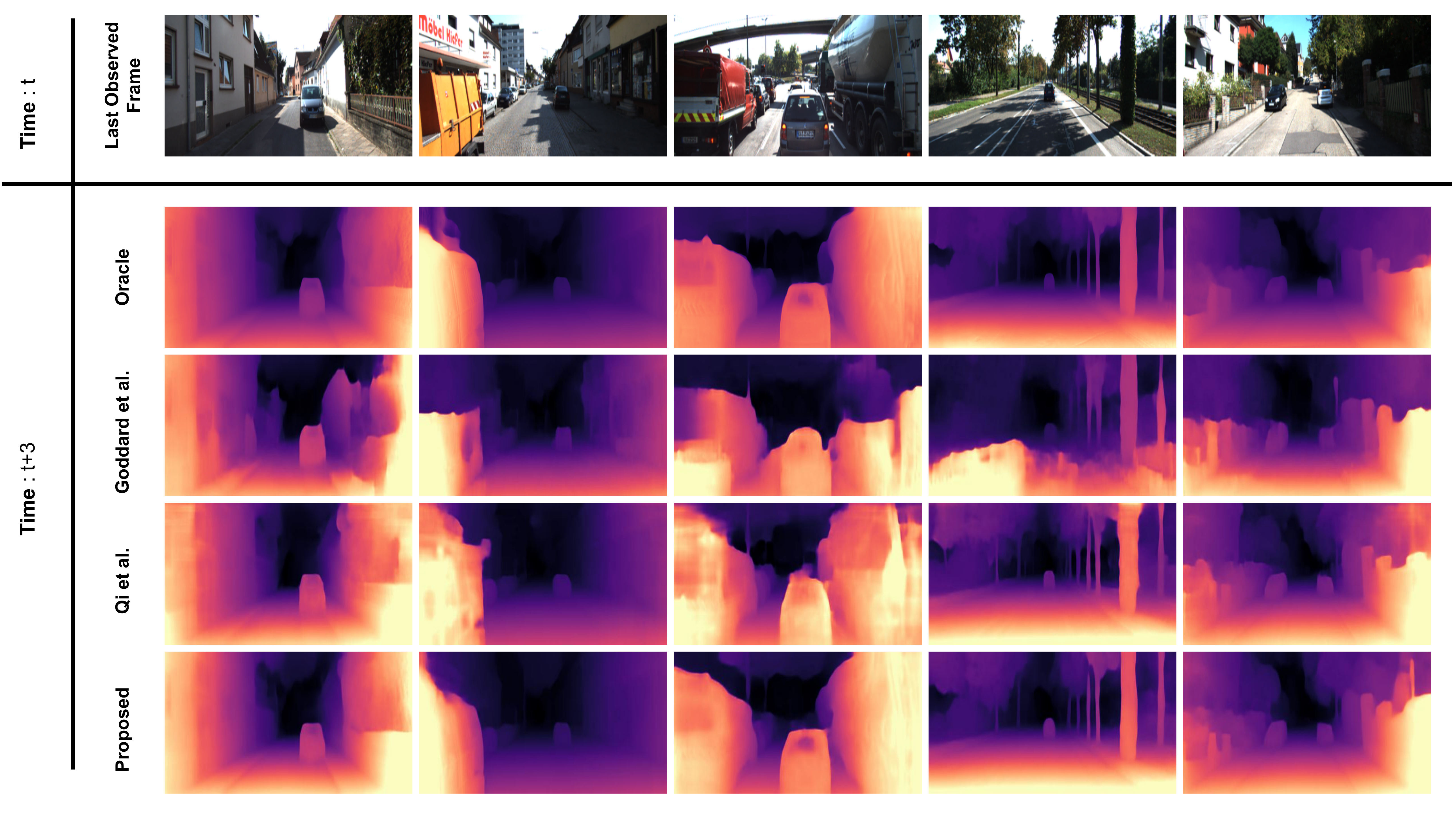}
\end{center}
\vspace{-.15in}
\caption{\textbf{Qualitative Illustration on KITTI:} Illustrates the short-term forecast on KITTI}
\vspace{-.22in}
\label{fig:archi}
\end{figure*}

\vspace{-0.15in}
\subsection{Ablation Studies}

\noindent \textbf{Effect of Channel Attention and Automasking in Ego-Motion} In order to validate the effectivity of the proposed PCAB block and automasking, we compare the test performance of our network trained without i) PCAB ii) Automasking iii) both PCAB and Automasking. From Tab 1, we can see that the variant (i) without PCAB drops in performance by $4/9\%$ in Abs Rel for short/mid forecast. This indicates that attention is necessary when there is a lack of annotated data. For variant (ii) without Automasking, we observe that the performance drop is slightly higher ($\sim 6/11\%$) in short/mid forecast. This is expected since inability to handle rigid motion is an existing problem even in self-supervised monocular depth setting. The illustration in Fig 4 clearly indicates the need of automasking for this task. For variant(iii), we observe the highest drop of $\sim 12/19\%$, which suggests that using a simple PoseCNN is ineffective for the task of depth forecasting similar to our previous finding.

\noindent \textbf{Does Ego-Motion Forecast help ?} In our approach we have forecasted rgb and flow to obtain a forecasted depth. The next natural question that can be asked is : \textit{Can we also forecast ego-motion to improve the forecasted depth ?}. In order to validate this setup, we adapted the ego-motion prediction module from \cite{qi20193d} in our forecasting setting denoted as $Ours^{\dagger}$. We obtained forecasted flow $\textbf{W}_{t+k}$, point cloud depth from our forecasted $\textbf{D}_{t+k}$,  and the pose estimate $[R|T]_{t-1,t}$ to predict $[R|T]_{t,t+k}$. From Tab 2, we can observe that the adapted setting ($Ours^{\dagger}$) performs competitively with our original setting for short term but surprisingly drops by $\sim 8\%$ in Abs Rel metric for mid-term forecast. A possible reason for this is that, the performance of ego-motion forecast is dependent on the quality of depth and flow maps. Since we are using forecasted depth and forecasted flow maps (which are generated in a unsupervised setting) to estimate the ego-motion forecast the performance drop in mid-term forecast is self-explanatory. This justifies our choice to not incorporate ego-motion forecast and instead condition the camera parameters on unobserved rgb frames.

\vspace{-0.10in}
\begin{figure}[!htbp]
\noindent\begin{minipage}{\linewidth}
  \noindent\begin{minipage}[b]{0.50\textwidth}
        \setlength{\tabcolsep}{12pt}
        \begin{tabular}{c|c|c}
        \hline
        \centering
        \multirow{2}{*}{Methodology} & Short & Mid \\ \cline{2-3} 
         & Abs Rel & Abs Rel \\ \hline
        Ours & $\textbf{0.107}$ & $\textbf{0.125}$ \\
        $Ours^{\dagger}$ & 0.108 & 0.134 \\ \hline
        \end{tabular}
        \vspace{0.2in}
         \newline Table 2: $\textbf{Effect of Ego-Motion forecast}$ on KITTI 
        dataset. We evaluate the effectiveness of Our approach vs our variant ($Ours^{\dagger}$) with ego-motion forecast on  short and mid level forecast
    \label{fblock}
  \end{minipage}%
  \hfill
  \begin{minipage}[b]{0.45\linewidth}
     \begin{figure}[H]
        \includegraphics[scale=0.12]{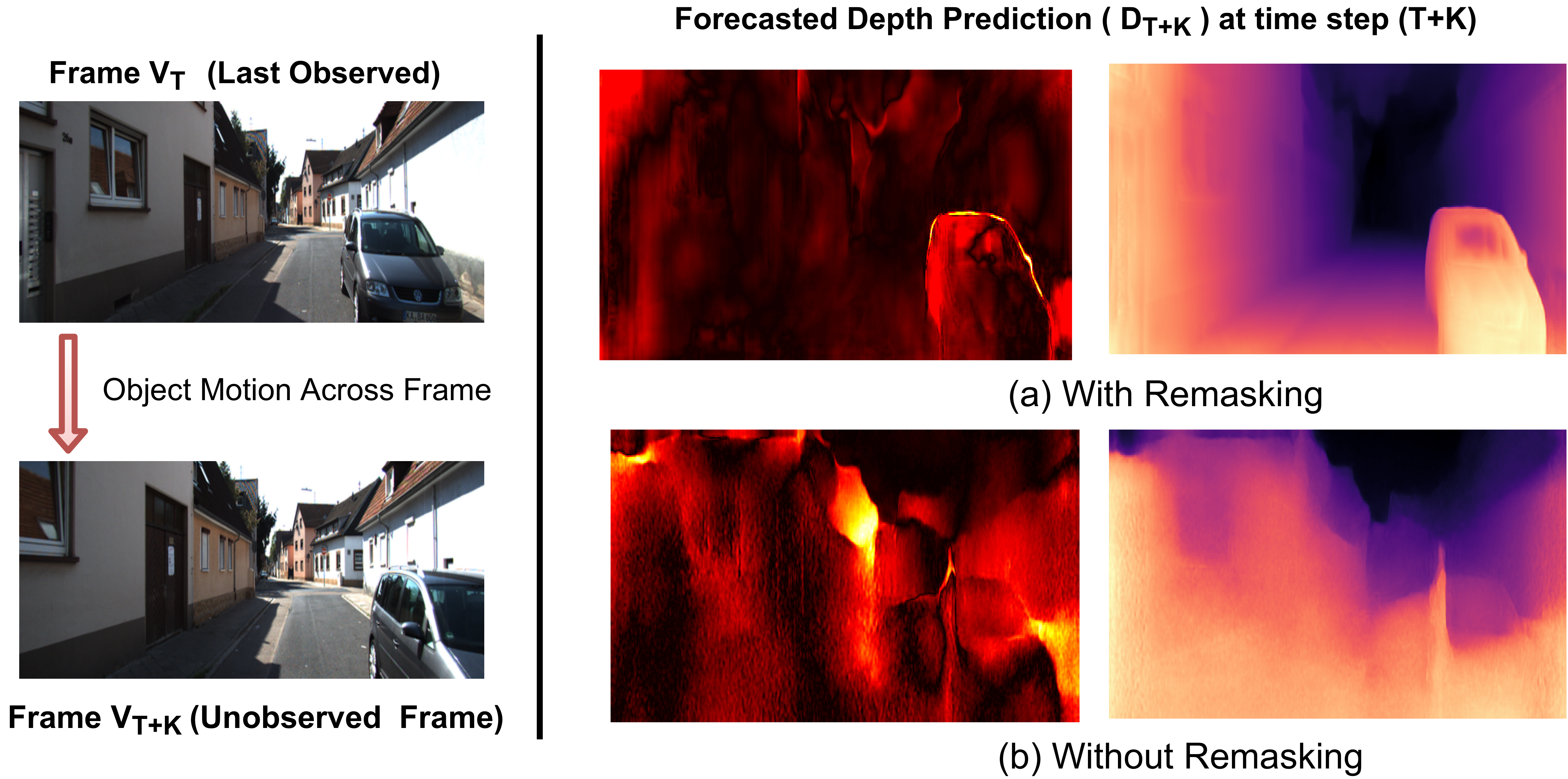}
        Figure 4. \textbf{Illustration of the effect of automask} on depth forecasting
        \label{fig:sample_figure}
    \end{figure}
    \end{minipage}%
  \end{minipage}
 \end{figure}
  \vspace{-0.10in}
\section{Limitations of our approach}
\vspace{-0.05in}
This problem although can work on wide array of unconstrained videos, it is not designed to handle jittery/unstable videos. As a result, our method will struggle to estimate depth for the videos captured using hand-held devices. Additionally, our approach is based on how the scene is illuminated -- which makes our design vulnerable to scene illumination. Hence, creating a design that is invariant to mode of capture and illumination is thus a part of our future work.
\section{Conclusion}
 \vspace{-0.10in}
We present a novel spatio-temporal forecasting framework DeFNet for the depth forecasting problem. Our method is designed to mitigate the practical barriers to predicting future frame depth such as annotation cost, generalization ability etc. We for the very first time solve the depth forecasting problem in a self-supervised fashion by formulating the depth estimation task as a novel-view synthesis problem. This allows our approach to solve the depth forecasting as an auxiliary task and thus making our approach annotation free. Experiments on two popular datasets verify the superiority of our approach. Moreover, being able to learn depth without the need for intrinsics/labels opens up the opportunity for pooling videos from any data sources together for a wider application.

\section{Acknowledgements}
We are grateful to the anonymous reviewers and also to Dr. Geethu Miriam Jacob, Dr. Xiatian Zhu for their insightful and constructive feedback on improving this work. 

\medskip


{
\small
\bibliographystyle{plain}
\bibliography{ref.bib}
}

\appendix

\end{document}